%% file: root.tex
\documentclass[letterpaper, 10 pt, conference]{ieeeconf}  % Comment this line out if you need a4paper

\IEEEoverridecommandlockouts                              % This command is only needed if 
\usepackage{amssymb}  % assumes amsmath package installed
\usepackage{pgfplots}

\usepgfplotslibrary{fillbetween,groupplots,dateplot}
\usetikzlibrary{patterns,shapes.arrows}
\pgfplotsset{
	max space between ticks=30pt,
	try min ticks=3,
	every axis/.style={
		axis y line=left,
		axis x line=bottom,
		axis line style={thick,->,>=latex, shorten >=-.4cm}
	},
	every axis plot/.append style={thick},
	tick style={black, thick},
	compat=newest
}
\tikzset{
	semithick/.style={line width=1.1pt},
}
\usepackage{booktabs}
\usepackage{multirow}
\usepackage{pifont}
\newcommand{\cmark}{\ding{51}}%
\newcommand{\xmark}{\ding{55}}%
\title{\LARGE \bf
Group Regression for Query Based Object Detection and Tracking
}

\author{Felicia Ruppel$^{1, 2}$, Florian Faion$^{1}$, Claudius Gl\"{a}ser$^{1}$ and Klaus Dietmayer$^{2}$% <-this % stops a space
\thanks{$^{1}$Robert Bosch GmbH, Corporate Research, 71272 Renningen, Germany, 
	{\tt\small \{firstname.lastname\}@de.bosch.com}}%
\thanks{$^{2}$Institute of Measurement, Control and Microtechnology, Ulm University, Germany,
	{\tt\small \{firstname.lastname\}@uni-ulm.de}}%
}

\usepackage{fancyhdr}
\usepackage{ragged2e}

\fancypagestyle{FirstPage}{
	
	\lfoot{\noindent\fontsize{7.5}{8.5}\selectfont\justifying\textcopyright\ 2023 IEEE.  Personal use of this material is permitted.  Permission from IEEE must be obtained for all other uses, in any current or future media, including reprinting/republishing this material for advertising or promotional purposes, creating new collective works, for resale or redistribution to servers or lists, or reuse of any copyrighted component of this work in other works.}
	
}
\makeatletter
\let\NAT@parse\undefined
\makeatother
\usepackage{hyperref}
\begin{document}

\maketitle
\thispagestyle{empty}
\pagestyle{empty}

%%%%%%%%%%%%%%%%%%%%%%%%%%%%%%%%%%%%%%%%%%%%%%%%%%%%%%%%%%%%%%%%%%%%%%%%%%%%%%%%
\begin{abstract}
	Group regression is commonly used in 3D object detection to predict box parameters of similar classes in a joint head, aiming to benefit from similarities while separating highly dissimilar classes. For query-based perception methods, this has, so far, not been feasible.
	We close this gap and present a method to incorporate multi-class group regression, especially designed for the 3D domain in the context of autonomous driving, into existing attention and query-based perception approaches. We enhance a transformer based joint object detection and tracking model with this approach, and thoroughly evaluate its behavior and performance. For group regression, the classes of the nuScenes dataset are divided into six groups of similar shape and prevalence, each being regressed by a dedicated head. We show that the proposed method is applicable to many existing transformer based perception approaches and can bring potential benefits. The behavior of query group regression is thoroughly analyzed in comparison to a unified regression head, e.g. in terms of class-switching behavior and distribution of the output parameters. The proposed method offers many possibilities for further research, such as in the direction of deep multi-hypotheses tracking.
\end{abstract}

%%%%%%%%%%%%%%%%%%%%%%%%%%%%%%%%%%%%%%%%%%%%%%%%%%%%%%%%%%%%%%%%%%%%%%%%%%%%%%%%
\section{Introduction}\label{sec_intro}
\thispagestyle{FirstPage}
3D object detection and tracking are essential components of a perception system for autonomous and automated driving. With the rise of deep learning for object detection in the 2D image domain \cite{NIPS2015_14bfa6bb}, building on this progress, methods for 3D object detection on point clouds were soon developed \cite{lang_pointpillars:_2019, Yang_2018_CVPR}. While many early approaches focused on transforming the point cloud so that it could be processed in 2D, it has recently been common practice to adapt these methods to the 3D domain explicitly \cite{zhu2019class} \cite{Reuse_2021_ICCV}. This includes augmentation strategies, such as ground truth (GT) sampling, but also model modifications, such as group regression heads: Object detection classes are divided into groups of classes, aiming to combine similar classes in shape, size and prevalence in the dataset, while segregating dissimilar ones to minimize their interference. The bounding boxes of grouped classes are regressed in a joint head \cite{zhu2019class}. The 3D domain-specific adaptations are implemented in major 3D object detection code bases such as MMDetection3D \cite{mmdet3d2020} and have shown to provide significant improvements in performance over the original version of many models, as can be found in the nuScenes [7] benchmark.

\begin{figure}[!t]
	\includegraphics[width=\columnwidth]{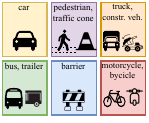}
	\caption{Illustration of classes that are combined into groups in the proposed group regression head for query based object detection and tracking. The grouping aims to separate dissimilar classes (in terms of prevalence and shape) while benefiting from in-group similarities.}
	\label{fig_title}
	\vspace{-5pt}
\end{figure}
Concurrently, another paradigm for object detection has been developed, which is based on transformer attention \cite{carion_end--end_2020}. Rather than relying on a dense grid of carefully placed anchor boxes, a set number of object queries is used \cite{misra_end--end_2021}. These queries interact with the input data via cross-attention in a transformer, aiming to obtain information about an object to detect. This approach was first developed for the image domain \cite{carion_end--end_2020} and then ported into 3D for detection and tracking on point clouds \cite{misra_end--end_2021, Bai_2022_CVPR, ruppel2022trans}. Some of them are suitable for autonomous driving and they rely on a 3D backbone \cite{Bai_2022_CVPR, ruppel2022trans}, which directly benefits from the aforementioned training augmentations for the 3D domain, especially GT sampling. However, to our knowledge, none of them allow to incorporate dedicated 3D group regression heads, but rather rely on a unified regression for all classes. Since group regression is so common with other 3D perception methods, this poses the question whether it should be adapted to query based approaches as well.

In this paper, we propose a group regression method applicable to many existing query perception approaches. Firstly, we analyze it in combination with a model from previous work, TransMOT \cite{ruppel2022trans}, which is a joint object detection and tracking approach. We enhance it with multi-class capability in two ways: Using a unified multi-class regression head as introduced for 2D \cite{carion_end--end_2020}, as well as comparing it to the proposed group regression heads, dedicated to 3D. Since our group regression can be applied to many existing transformer based object detection methods, we exemplarily show its application to TransFusion \cite{Bai_2022_CVPR}, which is a state-of-the-art object detector. We conclude whether it is beneficial to use such a head for multi-class detection and tracking. Besides this, we reflect on the commonly used augmentation of GT sampling and show whether it has a beneficial impact on query based detectors. The contributions of this paper are as follows:
\begin{itemize}
	\item Presentation of a novel method to incorporate multi-class group regression heads to transformer based object detection and tracking methods in 3D, as well as its combination with query refinement between layers.
	\item Analysis of the behavior of group regression compared to unified regression in an exemplary model from previous work \cite{ruppel2022trans}, as well as for TransFusion-L \cite{Bai_2022_CVPR}. Demonstration of the effect of disabling GT sampling for later training epochs.
	\item Conclusion under which circumstances a group regression is beneficial for query based perception.
\end{itemize}
This paper is structured as follows: In Section~\ref{sec_rel_work}, related work in object detection and tracking is summarized. In the following section, the proposed method is introduced. In Section~\ref{sec_results}, evaluation results are presented, mainly focusing on the difference between group regression heads and a unified regression, applied to two different models. Finally, a conclusion is drawn and an outlook is given.
\section{Related Work}\label{sec_rel_work}
In the context of autonomous driving, numerous approaches towards 3D perception been developed. Relevant related work for this paper can be divided into two fields: 3D object detection, especially anchor based as well as query based, and multi-object tracking.
\subsection{Anchor Based Object Detection}
The aim of 3D object detection is to estimate rotated 3D bounding boxes of surrounding objects in a certain sensor field of view. Sensor modalities commonly used for this are cameras \cite{Brazil_2019_ICCV,Shi_2020_CVPR}, lidar \cite{Yin_2021_CVPR,lang_pointpillars:_2019} and radar \cite{9827295}. We focus on lidar input in this paper, although the transformer architecture is flexible to accommodate for different input modalities \cite{pmlr-v164-wang22b}. As mentioned in the introduction, many 3D object detection methods are built on knowledge from the 2D domain. There, it is common to estimate objects relative to previously defined anchor boxes, which are placed at a certain location on the encoded feature map. In a two-stage approach, these anchor boxes are determined by a region proposal network \cite{NIPS2015_14bfa6bb}. In one-stage approaches, the anchor boxes are placed on a dense regular grid \cite{Lin_2017_ICCV, liu2016ssd}. The latter technique was adapted by many 3D object detectors. \cite{lang_pointpillars:_2019} and \cite{Yang_2018_CVPR} transform the input point cloud into a set of pillars or a projection into a 2D bird's-eye view plane, respectively. It is common to place an anchor of each class at each feature map location, often in two different, axis-aligned orientations \cite{Zhou_2018_CVPR}. Such anchors offer ideal conditions for group regression heads \cite{zhu2019class}: Because each anchor is pre-assigned to a class, it is processed only through the matching regression head. One shortcoming of such approaches is, however, the need for non-maximum suppression in post-processing \cite{lang_pointpillars:_2019}, as well as their restriction to dense grids.
\subsection{Query Based Object Detection}
With the success of transformers \cite{vaswani_attention_2017} in the natural language processing field, this concept has soon been adapted for computer vision \cite{khan2022transformers, carion_end--end_2020}. The first transformer object detector \cite{carion_end--end_2020} for images is anchor-free by design: A set of learned object queries, which are not bound to any location, is used to find objects in the image. For this, each of them interacts with the encoded input via cross-attention, aggregating features of interest, from which object bounding in absolute coordinate space on the image plane are regressed. However, additional research showed that it is beneficial to estimate bounding boxes relative to some anchor location, whether computed from learnt queries \cite{zhu2020deformable}, or used as initial prior to compute the object query encodings \cite{misra_end--end_2021}. So, current query based object detection methods cannot be classified as 'anchor-free'. However, they differ by only incorporating anchor locations, rather than axis-aligned anchor boxes. Besides this, queries are free to travel towards their location of interest, making it possible to work with significantly fewer queries than anchor boxes \cite{carion_end--end_2020}. Commonly, queries are not class-specific \cite{zhu2020deformable}. This makes it non-trivial to combine them with the mentioned 3D-specific group regression heads, which is the focus of this paper.

Concerning the 3D domain, transformer based detection solutions exist both with multi-view images as input \cite{pmlr-v164-wang22b}, as well as for lidar data \cite{misra_end--end_2021, Bai_2022_CVPR, ruppel2022trans}. This paper builds upon this line of work by proposing a universally applicable group detection head for such models, and by extending TransMOT \cite{ruppel2022trans} towards multi-class detection and tracking with group regression and unified regression.
\subsection{Query Based Multi-Object Tracking}
\begin{figure*}[!t]
	\vspace{5pt}
	\includegraphics[width=\textwidth]{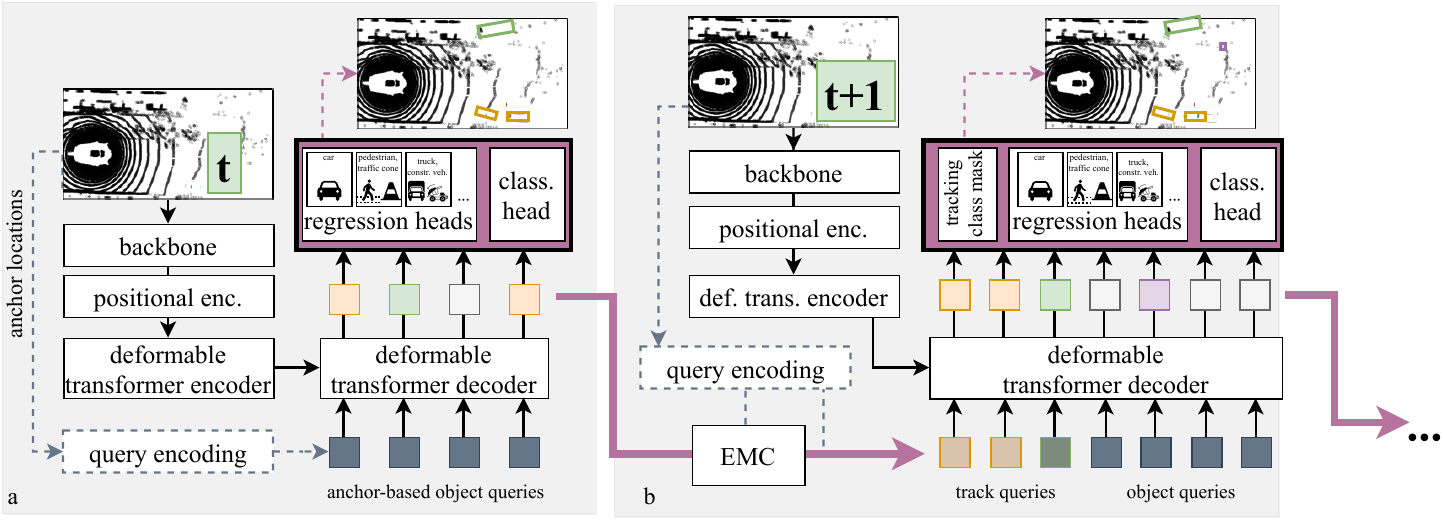}
	\caption{Overview of the proposed group multi-class regression method in combination with a joint object detection and tracking  model. On the left (a), an object detection step is pictured, which also serves as initialization for the tracker. The input point cloud is encoded through a backbone and a deformable transformer encoder. Besides this, anchor locations are sampled from it, which serve as priors for a set of object queries. They are passed through a deformable transformer decoder, accumulating information about objects to detect. Each of them can detect an object, or be assigned to the background class. To obtain bounding box parameters, each transformed query is passed into six regression heads, obtaining a box candidate for the respective group of classes, as well as one class head to obtain class scores. In tracking mode (right, b), feature vectors belonging to a detected object are propagated as track queries, aiming to continue this track. They are ego-motion corrected (EMC) by moving their anchor location accordingly.}
	\label{fig_model_overview}
\end{figure*}
Object queries collect information about an object in the transformer decoder in feature space. This makes them suitable to be propagated in time in order to carry track information to the following measured frame \cite{meinhardt_trackformer:_2021, ruppel2022trans}, allowing the tracker insight into the object's rich latent space, rather than relying on low-dimensional bounding boxes. These track queries are presented to the model in addition to the object queries, aiming to continue the track in its respective track query slot, so that the track association is solved by the transformer. Query based tracking methods are examples for a deep \textit{tracking-and-detection} paradigm. Contrary to this, classical trackers follow a \textit{tracking-by-detection} paradigm, such as \cite{chiu_probabilistic_2020}, which is based on a Bayesian filter. Some more sophisticated classical tracking approaches, such as the use of multiple hypotheses, are not common yet in the \textit{tracking-and-detection} paradigm on popular autonomous driving benchmarks \cite{ruppel2022trans}. In \cite{640267}, statistical tracking methods are discussed that maintain multiple possible states per tracked object, e.g. in terms of their class. The proposed group regression head in this paper opens the possibility to maintain multiple states per object as well, one per regression head. This is an interesting direction for further research, allowing to combine Bayesian principles with deep learning based trackers.
\section{Proposed Method}\label{sec_method}
We propose a multi-class grouping method that can be applied to many query based approaches. In the following, we first introduce its application to a particular method from previous work \cite{ruppel2022trans}, resulting in a transformer based multi-object tracking and detection multi-class model, which we call \textit{TransMOTM}. An overview of TransMOTM is pictured in Fig.~\ref{fig_model_overview}. Its components will be detailed in the following, as well as the application of group regression to other existing models.

\subsection{Tracking and Detection Model}
 In TransMOTM, transformer decoder queries serve as slots for newly detected objects (object queries) and continued tracks (track queries), following \cite{meinhardt_trackformer:_2021, ruppel2022trans}. The object queries are generated by sampling a selected number of point locations via farthest point sampling on the input point cloud that serve as anchor locations and applying a Fourier query encoding to them, as proposed by \cite{misra_end--end_2021}. These encodings are passed to the model as query positional encodings, so that they are added to the intermediate query states before each decoder layer. By design, each object query can detect one object of any class, which we denote as multi-potent queries. This capability is important to allow for a rather small total amount of queries, since the number of object queries needs to be larger than the maximum amount of expected objects in a frame. Alternatively, if there were specific queries for each class, the total number of queries would need to be larger than the sum of maximum expected numbers \textit{per class}. Multi-potent queries, however, are not directly compatible with multi-class group heads, which is detailed in \mbox{Section~\ref{sec_group_regression}}.

TransMOTM can be run in detection mode, which is pictured in Fig. \ref{fig_model_overview}a.
The measured point cloud is encoded by a 3D backbone, such as PointPillars \cite{lang_pointpillars:_2019}. The resulting feature map is input to a transformer that is based on deformable attention \cite{zhu2020deformable}. In each of six decoder layers, the object queries can interact with one another through self-attention, followed by interaction with the input feature map through deformable cross-attention. At any point in the pipeline, a regression head can be applied to a query to obtain the bounding box it encodes (or bounding boxes in the case of a multi-group head), making them interpretable as a feature vector belonging to a location in 3D space.

This makes these feature vectors suitable to encode a track state \cite{ruppel2022trans, meinhardt_trackformer:_2021} in the model's tracking mode, which is pictured in Fig. \ref{fig_model_overview}b. The feature vectors belonging to detected objects at timestep $t$ are propagated to the next timestep and offered as track queries to the model, besides the object queries. The model is trained to update the track queries and continue the respective track in the following frame. Once a feature vector is propagated as a track query it can not change its class, which is ensured by a tracking class mask. In autonomous driving, it is common for an ego vehicle to move significant distances between two frames, especially considering a dataset such as nuScenes \cite{caesar_nuscenes:_2020}, which has annotated frames at a rate of $2$ Hz. To correct the ego-motion, for each track query, a new anchor location is computed and encoded, which places the object in the new measurement's coordinate frame and is passed to the model as positional encoding in addition to the track queries.

To obtain a detected or tracked bounding box from a feature vector, a regression head is applied to it. The proposed group multi-class regression heads in contrast to unified regression heads are detailed in the following.
\subsection{Unified Multi-Class Regression Head} 
A unified multi-class regression head is a multi-layer perceptron (MLP) that decodes a feature vector into a low dimensional bounding box of any class with the parameters
\begin{equation}
b=(\Delta x,\Delta y,\Delta z,w,l,h,\textrm{sin}(\gamma), \textrm{cos}(\gamma), v_x,v_y),
\end{equation}
denoting the position, size, orientation and velocity of an object relative to their anchor location. Similar unified heads are commonly used in query based object detection, both in the image domain \cite{carion_end--end_2020} as well as in 3D \cite{Bai_2022_CVPR}. Which class an object belongs to is estimated by a second MLP, the class head, that outputs a classification score for each class, as well as the 'no-object' class, which describes background, following \cite{carion_end--end_2020}. During object detection training, each query needs to be assigned either to a GT object or to the background class based on its estimated bounding box and class score, to then compute a loss. For this, a cost matrix $C_U\in\mathbb{R}^{T\times M}$ is computed, comparing each of the $T$ ground truth (GT) boxes to every of $M$ estimated boxes. The Hungarian algorithm is used to match each GT box to exactly one estimate, minimizing the total cost. During tracker training, track queries are matched to the respective continued tracks, while object queries can accommodate for newly spawned tracks and are matched via Hungarian matching as described before.
\subsection{Group Multi-Class Regression}\label{sec_group_regression}
As mentioned in the introduction, group regression heads are a common practice with many anchor based object detection models \cite{zhu2019class}. A group regression suitable for query based perception methods is proposed in the following.  With anchor based methods, it is common to use class-specific anchors, making it trivial to pass each anchor to the matching head. However, we want to facilitate multi-potent queries, i.e. that each of them can become any class. This needs to be accommodated both during training and inference. To partition the nuScenes \cite{caesar_nuscenes:_2020} classes into groups, we follow~\cite{zhu2019class}, obtaining six classes: (car), (truck, construction vehicle), (bus, trailer), (barrier), (motorcycle, bicycle), (pedestrian, traffic cone), as illustrated in Fig. \ref{fig_title}. Therefore, six MLPs are trained, each decoding a feature vector into a set of bounding box parameters $b_i$, $i=1,\dots, 6$. During detector training, the estimated boxes need to be matched to GT boxes, but, even though there are now six box candidates per query rather than only one, the computational effort for matching shall not increase. Therefore, a cost matrix $C_G\in\mathbb{R}^{T\times M}$ is generated that has the same size as with a unified class head, one line per GT box and one column per query. For each line $t$ in $C_G$, the respective GT box $b^{\left[t\right]}$ of class $c^{\left[t\right]}$ is compared to each query's estimated box $b^m_{g_{c^{\left[t\right]}}}$, $m=1,\dots, M$, where $g_{c^{\left[t\right]}}$ selects the group head that matches the GT object's class, $c^{\left[t\right]}$. This ensures that each query can be matched to any class, while guaranteeing that it is not matched to multiple classes, since any estimated box is matched at most to one GT box. The regression loss is computed between each GT box and its match, ensuring that this loss only affects the respective group head, while leaving the others untouched. This allows each group head to specialize on their assigned classes.
\subsection{Query Refinement}
Between decoder layers, the queries' anchor locations are refined to move them closer to the object they are aiming to estimate, following \cite{zhu2020deformable, ruppel2022det}. With group regression, this poses the question which of the six estimated object locations per query to use for refinement. Two options are depicted in Fig. \ref{fig_refinement_illustration}: One could either move towards the strongest class (B), or compute a weighted mean of all available bounding box location estimates. For this, the class scores belonging to the same group are summed, resulting in group scores, which are used as weights. To avoid gradient instability during training, option A is used. Whether it is beneficial to adopt option B during inference is evaluated in the results section. 
\subsection{Application of Group Regression to Related Methods}
It is feasible to apply the proposed group regression to many existing query based perception approaches, both to methods relying on lidar input \cite{Bai_2022_CVPR}, as well as image input \cite{pmlr-v164-wang22b, jiang2022polarformer, li2022bevformer}. Exemplary, we apply it to the state-of-the-art detector TransFusion \cite{Bai_2022_CVPR}, which is detailed in the results section. What all these query based perception methods have in common is a similar set-based loss computation pipeline that was first introduced by \cite{carion_end--end_2020}: A set of queries is transformed into a set of feature vectors containing information about objects to detect. Through a regression head, each of them constitutes a bounding box candidate. Via matching, an assignment between the candidates and GT is computed. In this regard, they are similar to what we described for TransMOTM. Therefore, the adapted matching strategy compatible with group regression can conveniently be applied to them as well.
\subsection{Fade-out Strategy}
As introduced in Section \ref{sec_intro}, 3D object detection methods benefit from data augmentation techniques, one of which is GT sampling. This entails randomly placing objects of underrepresented classes into measured frames. While this technique certainly improves the prevalence of positive examples for such classes and therefore improves the performance \cite{zhu2019class}, it poses the question whether such 'unnatural' placement of objects can harm the model's understanding of the data distribution. This is especially crucial in a model, such as those in the scope of this paper, that can reason on a more global scale through cross attention, rather than only incorporating local features for object detection \cite{lang_pointpillars:_2019}. To investigate this, we adopt a fade-out strategy as introduced in \cite{Wang_2021_CVPR}: After training for 35 epochs with GT sampling, the GT sampling is disabled for another five epochs, aiming to present the model a data distribution that is more similar to the one during testing and for real-world usage.
\begin{figure}[!t]
	\includegraphics[width=\columnwidth]{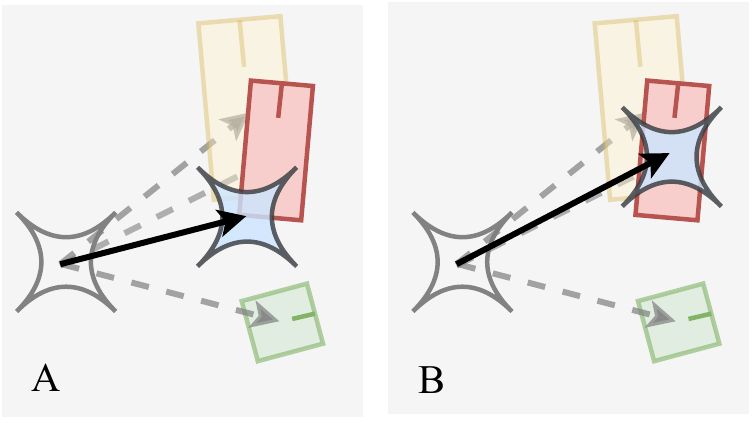}
	\caption{Query refinement illustration. The colored boxes depict multiple bounding box hypotheses that stem from one query. A: The query is moved to a location obtained with a weighted mean between group estimates. B: The query is moved to the position of the most prominent class.}
	\label{fig_refinement_illustration}
\end{figure}
\section{Results}\label{sec_results}
The focus of this section is to investigate the effect of using group regression with query based perception, both in terms of model performance as well as regression head behavior. For this, a thourough evaluation is first carried out with TransMOTM, the model presented in this paper. Since a more basic backbone, PointPillars \cite{lang_pointpillars:_2019} is used, it is not expected to reach top performance on a benchmark. However, the comparison between unified and group regression still applies. Besides this, the proposed group regression is applied to and evaluated with TransFusion-L \cite{Bai_2022_CVPR}, a state-of-the-art detector with a more expressive backbone.
\subsection{Group Regression With Object Detection}
Results for object detection on the nuScenes dataset of different versions of TransMOTM with and without group regression as well as with and without the fade strategy are listed in Table \ref{tab:detection_results}. The experiments are carried out on the nuScenes {\tt val} dataset. The metrics show that, while a group regression head is beneficial for some classes and increases the mean AP score, its performance is very similar to a unified regression head. This implies that this query based 3D detection model is able to compensate for class differences in shape and distribution on the nuScenes dataset, even with a single regression head. On the other hand, group regression does not harm the performance, while slightly improving some classes' metrics. Therefore, they can safely be used in combination with certain use-cases, e.g. in the case of a stronger shape imbalance between classes, or for multi-hypothesis detection, as detailed in the following subsections.

\begin{table*}[t]
	\setlength\tabcolsep{0pt} % make LaTeX figure out intercolumn spacing
	\caption{Object Detection with model variants}
	\label{tab:detection_results}
	\begin{tabular*}{\textwidth}{@{\extracolsep{\fill}} ll cccccccc}
		\toprule
		Method & Regr. type & Fade&  mAP$\uparrow$ & AP (car)$\uparrow$& AP (truck)$\uparrow$&AP (pedestrian)$\uparrow$ &AP (bus)$\uparrow$& AP (bicycle)$\uparrow$&AP (trailer)$\uparrow$ \\
		\midrule
		TransMOT \cite{ruppel2022trans} & unified &-&-&
		0.727& -&-&-&-&-\\
		\midrule
		\multirow{4}{*}{TransMOTM} & unified & \xmark &0.329&0.734&0.419&0.498&0.578&0.042&0.271 \\
		&group & \xmark & 0.339&0.736&0.436&0.500&0.589&0.046&0.273\\
		&unified & \cmark &0.380&0.755& 0.453&\textbf{0.541}&\textbf{0.603}&0.116&0.320\\
		&group & \cmark&\textbf{0.384}& \textbf{0.758}&\textbf{0.455}&0.539&0.590&\textbf{0.135}&\textbf{0.326}   \\
		\bottomrule
		
	\end{tabular*}
	%\vspace{-10pt}
\end{table*}
\subsection{Inter- and Intra-Group Regression Head Variance}
In the case of multi-group regression heads, one would expect that the model learns the distribution of shapes for each regression head. Therefore, the variance between output values of the same head should be small compared to the total variance of all heads, since classes are grouped according to their appearance. To quantify this, estimated bounding boxes are collected for each group head over all frames of the nuScenes {\tt val} dataset. The result of this is pictured in Fig.~\ref{fig_intra_wlh}, exemplary for the length parameter. It can be observed that the regression heads operate in ranges that were to be expected, considering the classes they contain, i.e. that they are able to learn the respective shape distribution.

The variance between different heads on the other hand may be larger, since different kinds of objects are estimated at the same time. In this regard, we evaluate each query individually: A query results in six object candidates. How do they vary, compared to one another, on a per-query basis? For this evaluation, a sample standard deviation is computed for each object query in each frame and decoder layer between the six regression heads' outputs. The medians of these values are depicted in Fig. \ref{fig_inter_group_var} for the location parameters $x$, $y$ and $z$, along with their $25$th and $75$th percentiles. It is observable that the $x$ and $y$ parameters vary more between heads than $z$. This means that the model does not always place the six bounding box candidates per query in the same spot, but that they can deviate. This supports the understanding that they can be interpreted as multiple object hypotheses. It is, however, notable that queries tend to have similar regions of interest through all their regression heads, i.e. the inter-group standard deviation does not span the entire sensor range of $100$m, but a smaller excerpt of it.
\begin{figure}[t]
	\vspace{10pt}
	\input{figures/inter_location}
	\caption{Each query results in six bounding box hypotheses through the six group regression heads. Between those, a standard deviation of the location parameters $x$, $y$ and $z$ is computed. The lines show the median of the standard deviations obtained from each query and decoder layer over the nuScenes {\tt val} dataset. The colored areas depict the $25$th and $75$th percentiles, respectively.}
	\label{fig_inter_group_var}
\end{figure}
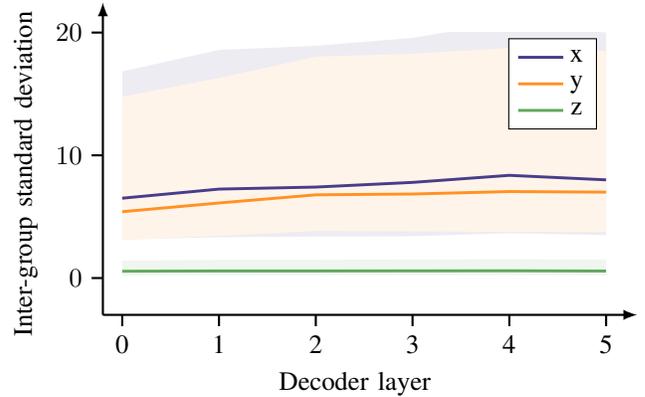
\begin{figure}[t]
	\vspace{10pt}
	\input{figures/intra_wlh}
	\caption{Median object length parameter as estimated by different group regression heads on the nuScenes {\tt val} dataset, with $25$th and $75$th percentiles as error bars. It can be observed that the model is able to learn the shape distribution of the respective groups. Notation: \textit{car:} car, \textit{tr./con.:} truck, construction vehicle, \textit{bus/tr.:} bus, trailer, \textit{barrier:} barrier, \textit{bike:} motorcycle, bicycle, \textit{ped./co.:} pedestrian, traffic cone}
	\label{fig_intra_wlh}
\end{figure}
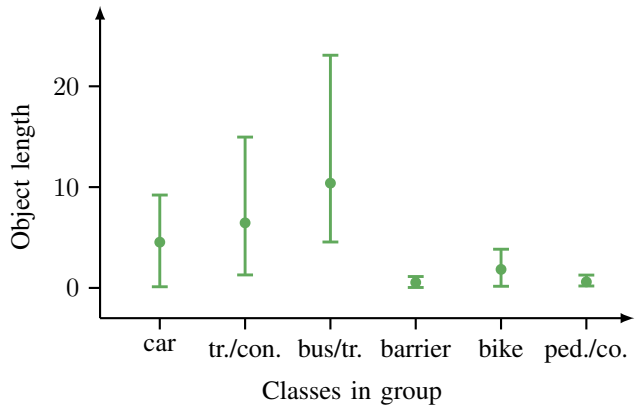
\subsection{Refinement analysis}
As mentioned above, the placement of refined anchor locations between decoder layers is a design choice, especially during inference. They can be placed at the mean of the heads' output locations weighted by the respective group scores, as we do during training, which is pictured in Fig. \ref{fig_refinement_illustration}(A). Alternatively, it may be beneficial to place them at the strongest classes location, illustrated in Fig. \ref{fig_refinement_illustration}(B). Both versions are compared, the results of which can be found in Table \ref{tab:refinement_ablation}. We find that switching to a top-candidate mode during inference harms the performance. This may be explained by the different setup compared to the training, as well as the possibility for class switches, i.e. a change of the most prominent class between decoder layers..
\begin{table}[h]
	\setlength\tabcolsep{0pt} % make LaTeX figure out intercolumn spacing
	\caption{Refinement ablation}
	\label{tab:refinement_ablation}
	\begin{tabular*}{\columnwidth}{@{\extracolsep{\fill}} ll cccccc}
		\toprule
		Method  & mAP$\uparrow$ & mATE (m)$\downarrow$& mASE (1-IOU) $\downarrow$\\
		\midrule
		Weighed Mean &\textbf{0.384} &
		\textbf{0.534}& 0.264\\
		Top Candidate   &0.378&	0.538&	0.264 \\
		\bottomrule
		
	\end{tabular*}
	%\vspace{-10pt}
\end{table}
\subsection{Fade-out Strategy}
As can be seen in Table \ref{tab:detection_results}, the fade-out strategy significantly improves the performance throughout all classes. This can be explained by a reduced number of false positives due to the model's insight into a more realistic data distribution during the fade-out phase. From these observations, we can conclude that this fade-out strategy should be more widely adopted, especially for detection methods that operate on a larger spatial context for detecting an object rather than a local one.
\subsection{Class Switches Between Layers} \label{sec_cls_switch}
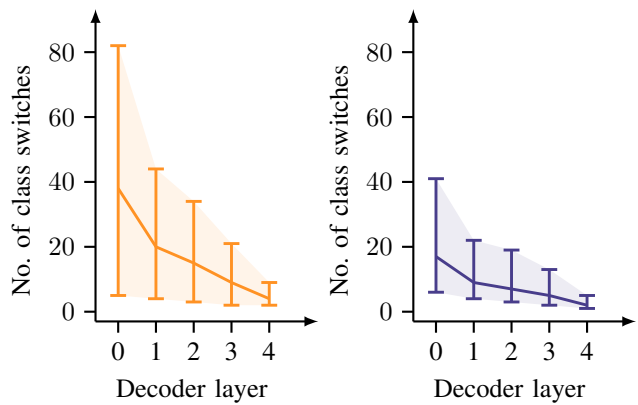
\begin{figure}[t]
	\vspace{10pt}
	\input{figures/class_switches_nofade}\hspace{5pt}
	\input{figures/class_switches_fade}
	\caption{Median observed class switches after each decoder layer on the nuScenes {\tt val} dataset of TransMOTM with group regression. The error bars denote $25$th and $75$th percentiles. \textit{Orange:} without fade out training, \textit{blue:} after fade out.}
	\label{fig_class_switches}
\end{figure}

In the proposed detection approach, the queries are not class-restricted, i.e. they can switch classes anytime through the layers of the model. We evaluate how often such class switches occur. In the following, we define a class switch as a change of the class with the highest score between two consecutive layers. The 'no-object' class is also considered a class. Class switches are counted on the nuScenes {\tt val} dataset. We observe that class switches occur similarly often in the group regression and unified regression models. We find, however, that the fade-out strategy has a significant impact on the class switch behavior. As can be seen in Fig.~\ref{fig_class_switches}, with a group regression model trained without fade-out on the left, as well as one with fade-out on the right, the latter exhibits circa half as many class switches. Further analysis shows that especially switches between the 'no-object' class and another class and vice-versa are reduced. This implies that the fade-out strategy improves the model performance by reducing the occurrence of doubtful cases, which could either be an object or background. This is in line with the observation that the fade-out strategy reduces the number of false positive detections. 
\begin{table*}[t]
	\setlength\tabcolsep{0pt} % make LaTeX figure out intercolumn spacing
	\caption{Joint tracking and detection results}
	\label{tab:tracking_results}
	\begin{tabular*}{\textwidth}{@{\extracolsep{\fill}} ll cccccccccc}
		\toprule
		Method & Regr. type & Fade& AMOTA$\uparrow$ &AMOTP$\downarrow$& AMOTA (car)$\uparrow$&AMOTA (truck)$\uparrow$&AMOTA (pedestrian)$\uparrow$&AMOTA (bus)$\uparrow$ \\
		\midrule
		TransMOT \cite{ruppel2022trans} & unified & -& -&-&0.674&-&-&- \\
		\midrule
		\multirow{2}{*}{TransMOTM (ours)} &unified & \cmark &0.378& 1.281&0.662&\textbf{0.385}&0.331&0.623\\
		&group & \cmark &\textbf{0.385}&\textbf{1.245}&\textbf{0.680}& 0.383&\textbf{0.333}&0.623\\
		\bottomrule
		
	\end{tabular*}
	%\vspace{-10pt}
\end{table*}
\begin{table*}[t]
	\setlength\tabcolsep{0pt} % make LaTeX figure out intercolumn spacing
	\caption{Comparison of the proposed group head applied to TransFusion-L \cite{Bai_2022_CVPR}}
	\label{tab:detection_results_transfusion}
	\begin{tabular*}{\textwidth}{@{\extracolsep{\fill}} ll cccccccc}
		\toprule
		Method & Regr. type & Fade&  mAP$\uparrow$ & AP (car)$\uparrow$& AP (truck)$\uparrow$&AP (pedestrian)$\uparrow$ &AP (bus)$\uparrow$& AP (bicycle)$\uparrow$&AP (trailer)$\uparrow$ \\
		\midrule
		TransFusion-L \cite{Bai_2022_CVPR} & unified & \cmark &\textbf{0.611}&0.851&0.552&\textbf{0.864}&0.692&\textbf{0.492}&\textbf{0.404} \\
		TransFusion-L - grouped &group & \cmark & 0.600&\textbf{0.854}&\textbf{0.554}&0.855&0.680&0.457&0.396\\
		TransFusion-L - single&singular & \cmark &0.602&0.849& 0.538&0.857&\textbf{0.693}&0.470&0.395\\
		\bottomrule
		
	\end{tabular*}
	%\vspace{-10pt}
\end{table*}
\subsection{Multi-Hypotheses in Detection}
\begin{figure}[t]
	\includegraphics[width=\columnwidth]{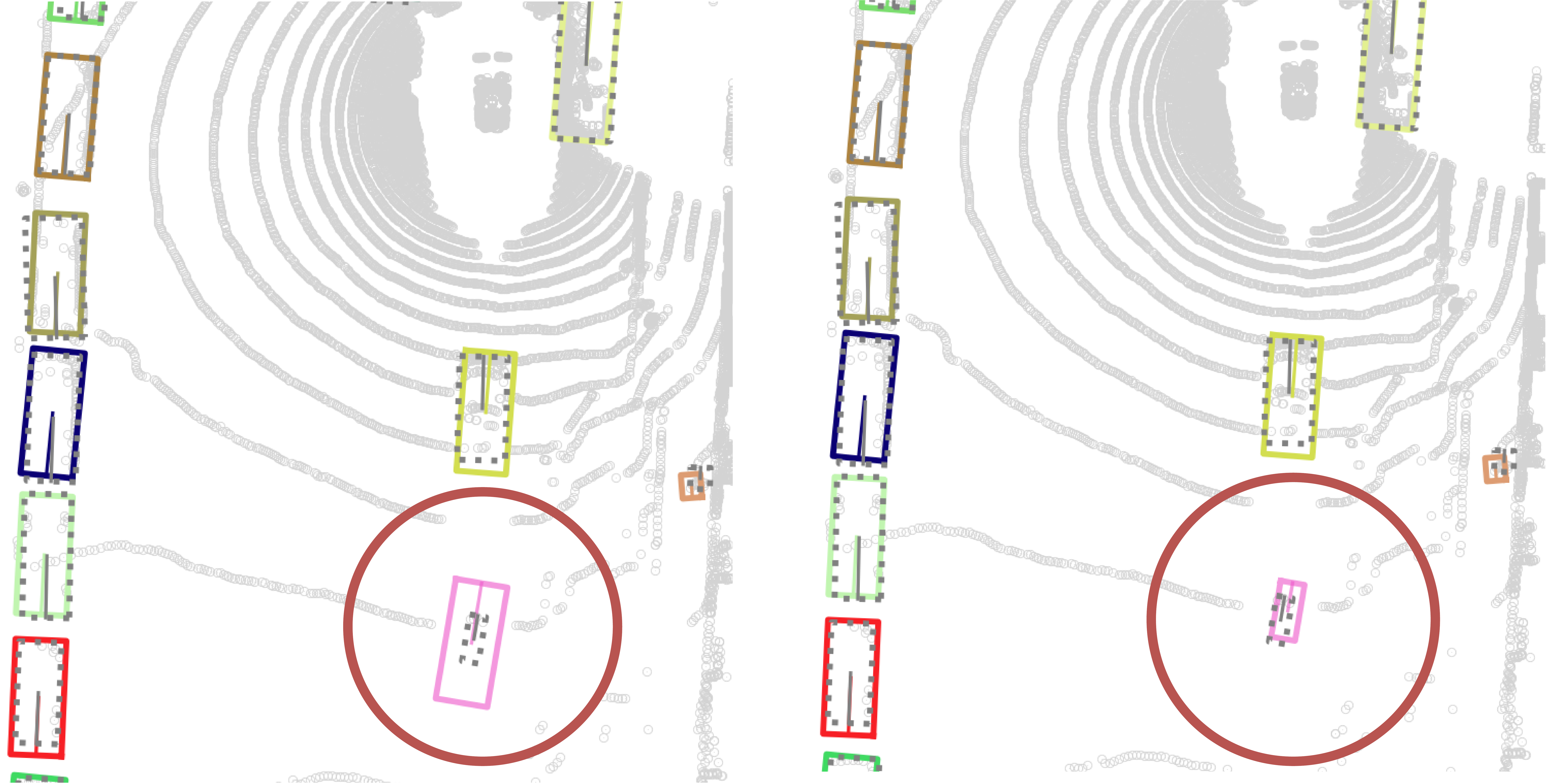}
	\caption{Object detection example, excerpt from the full birds-eye-view. Ground truth objects are marked in dotted lines, detections in colored boxes. \textit{Left:} The bounding boxes with strongest class scores are pictured. The circled bicycle is mistaken for a car. \textit{Right:} The circled bounding box was replaced by the second strongest hypothesis, the bicycle/motorcycle group head output. Its assigned group score is $0.19$}
	\label{fig_multi_hypotheses}
\end{figure}

Through the six group regression heads, each query produces six bounding box candidates. These can be interpreted as multiple object hypotheses, each with their respective group score, i.e. the sum of the contained classes' scores. An example for this is given in Fig.~\ref{fig_multi_hypotheses}: On the left side, the detector's output is depicted, only showing the most prominent class per query. A bicycle is mistaken for a car (circled red). If, however, the second most prominent hypothesis of this query is taken into account, the bycicle can be detected (right image). This can be beneficial in two ways: For object detection, multiple hypotheses with their respective scores can be output and passed to downstream modules, allowing them to make use of this additional information. This is especially interesting if uncertainty is involved, i.e. if the model is unsure which the correct hypothesis is. For tracking, multiple hypotheses allow a track id to stay consistent, even if it is unclear at track initialization which class an object belongs to. This means that a track could be initialized with a track query that maps to multiple possible bounding boxes, which can then be refined in the following frames, when the object is observed again. This would allow a switch of the most prominent class, i.e. the strongest hypotheses over the course of the track without needing to initialize a new track. This concept is similar to multiple model methods for tracking \cite{640267}, which would then be applicable to deep tracking with transformer based methods, which is an interesting direction for further research.

\subsection{Tracking Results}
Results for the tracking performance of the proposed model can be found in Table \ref{tab:tracking_results}. The tracker is trained in a two-step approach: First a detection model is trained. Then the tracker is trained on frame pairs for another 15 epochs. Fade indicates the usage of the fade-out strategy during detector training, while GT sampling is deactivated during tracker training. The proposed method outperforms TransMOT \cite{ruppel2022trans}, which was car-specific and has shown to be superior to a Kalman-Filter based baseline on given transformer detections \cite{ruppel2022trans}. Concerning the comparison between the variants with and without group regression heads, we find that employing class regression heads slightly outperforms the alternative. Especially with follow-up goals in mind, such as a multiple model methods \cite{640267} group regression heads in combination with query based tracking can be a way to make this possible.

\subsection{Application to Existing Methods}
Exemplary, the proposed grouping strategy is applied to TransFusion-L \cite{Bai_2022_CVPR}, a query based state-of-the art object detection method. This approach constitutes a Voxelnet \cite{Zhou_2018_CVPR} backbone with voxels of size $\left[0.075, 0.075, 0.2\right]$ meters, which is much more expressive than what the model we evaluated in the previous subsections, TransMOTM, incorporated. Besides this, TransFusion entails an informed query initialization strategy that places the initial query locations at probable object locations as determined from the Voxelnet feature map. The transformer decoder's task therefore is a bit different in this context: While it needs to find objects from randomly sampled query anchor locations in TransMOTM, it only refines promising object locations in TransFusion.

To incorporate group regression, we modify \mbox{TransFusion-L} to entail six different regression heads, one for each group of classes and train according to the proposed method. Besides this, we evaluate a second version that contains $10$ groups, i.e. one group per class. The matching strategy to ground truth objects is the same as described in the main body of this paper. Both the grouped and non-grouped versions of Transfusion are trained with a pretrained backbone, for $5$ epochs with GT augmentation, as well as $5$ epochs without. The results are reported in Table~\ref{tab:detection_results_transfusion}. Even though the transformer is used in a different way in this model, we find a similar behavior as TransMOTM. i.e. that some classes perform better in one variant and some in the other.

\section{Conclusion}\label{sec_concl}
In this paper, a method for multi-class group regression heads, which is applicable to many query based perception methods, was presented. We were able to show that the proposed method can slightly improve performance compared to a unified head, both for object detection as well as tracking, with TransMOTM. However, we find that it is not necessary to use group regression with the standard nuScenes classes, as a unified head achieves similar performance. This is surprising, since multi-group regression is very common with most 3D object detection methods \cite{mmdet3d2020}.  It is to be evaluated whether some of them could use unified heads without compromising performance as well \cite{peri2023towards}. However, we showed that query group regression is able to learn the grouped object's shape distributions and that it does not harm overall performance, making it applicable to the following use-cases: The shape imbalance between classes of interest could be larger than those in nuScenes, e.g. when also detecting elongated road boundaries. One could use group regression to parameterize such a class differently than the others and to learn its shape. Besides this, we offer the option to perform multi-hypotheses detection and tracking, as detailed above. This suggests interesting directions for further research, both in the direction of handling uncertainty in downstream modules, as well as towards multiple model methods \cite{640267} for deep tracking.
\addtolength{\textheight}{-2cm}   % This command serves to balance the column lengths
\bibliographystyle{IEEEtran}

\bibliography{references}

\end{document}

%% file: figures/inter_location.tex
% This file was created with tikzplotlib v0.9.12.
\begin{tikzpicture}

\definecolor{color0}{rgb}{0.2823529411764706, 0.23921568627450981, 0.54509892156862}
\definecolor{color1}{rgb}{1.0, 0.5725490196078431, 0.1411764705882353}
\definecolor{color2}{rgb}{0.3803921568627451, 0.6627450980392157, 0.3607843137254902}

\begin{axis}[
height=151.86214363138544,
tick align=outside,
tick pos=left,
width=235.71811,
x grid style={white!69.0196078431373!black},
xmin=-0.1999999959021806, xmax=5.010000024996698,
xlabel={Decoder layer},
ylabel={Inter-group standard deviation},
xtick style={color=black},
xtick={0,1,2,3,4,5},
xticklabels={0,1,2,3,4,5},
y grid style={white!69.0196078431373!black},
ymax=20,
ymin=-3,
%ymode=log,
ytick style={color=black}
]

\addplot [semithick, color0]
table {%
	0 6.50502705574036
	1 7.24769735336304
	2 7.41196513175964
	3 7.7927713394165
	4 8.36829519271851
	5 8
};

\addplot [semithick, color1]
table {%
	0 5.40492129325867
	1 6.1166832447052
	2 6.78027868270874
	3 6.84696054458618
	4 7.04724645614624
	5 7
};
\addplot [semithick, color2]
table {%
	0 0.555234432220459
	1 0.576831102371216
	2 0.576419860124588
	3 0.58415499329567
	4 0.59123820066452
	5 0.57
};

\addlegendentry{x}
\addlegendentry{y}
\addlegendentry{z}
\addplot [name path=upper0b,draw=none] table{%
	0 82
	1 44
	2 34
	3 21
	4 9
};
\addplot [name path=lower0,draw=none]
table {%
	0 3.15369713306427
	1 3.32387697696686
	2 3.37983667850494
	3 3.40916752815247
	4 3.65251457691193
	5 3.5
};
\addplot [name path=upper0,draw=none]
table {%
	0 16.8318316936493
	1 18.5854208469391
	2 18.9184463024139
	3 19.5650718212128
	4 20.8825495243073
	5 20
};
\addplot [fill=color0!10] fill between[of=upper0 and lower0];

\addplot [name path=lower1,draw=none]
table {%
	0 3.08642822504044
	1 3.44060027599335
	2 3.83081996440887
	3 3.80871516466141
	4 3.75299602746964
	5 3.75
};
\addplot [name path=upper1,draw=none]
table {%
	0 14.7776646614075
	1 16.3076062202454
	2 18.0420229434967
	3 18.2768964767456
	4 18.7502303123474
	5 18.5
};
\addplot [fill=color1!10] fill between[of=upper1 and lower1];

\addplot [name path=lower2,draw=none]
table {%
	0 0.203810691833496
	1 0.23736435174942
	2 0.233141697943211
	3 0.233595252037048
	4 0.236536413431168
	5 0.23
};
\addplot [name path=upper2,draw=none]
table {%
	0 1.41044333577156
	1 1.48110434412956
	2 1.48542442917824
	3 1.50618353486061
	4 1.52600748836994
	5 1.5
};
\addplot [fill=color2!10] fill between[of=upper2 and lower2];
\end{axis}
\vspace{10pt}
\end{tikzpicture}

%% file: figures/intra_wlh.tex
\newcommand{\boundellipse}[3]% center, xdim, ydim
{(#1) ellipse (#2 and #3)
}% This file was created with tikzplotlib v0.9.12.
\begin{tikzpicture}

\definecolor{color0}{rgb}{0.2823529411764706, 0.23921568627450981, 0.54509892156862}
\definecolor{color1}{rgb}{1.0, 0.5725490196078431, 0.1411764705882353}
\definecolor{color2}{rgb}{0.3803921568627451, 0.6627450980392157, 0.3607843137254902}

\begin{axis}[
height=151.86214363138544,
tick align=outside,
tick pos=left,
width=235.71811,
x grid style={white!69.0196078431373!black},
xmin=-0.1999999959021806, xmax=5.710000024996698,
xlabel={Classes in group},
ylabel={Object length},
xtick style={color=black},
xtick={0.5,1.5,2.5,3.5,4.5,5.5},
xticklabels={car,
	tr./con.,
	bus/tr.,
	barrier,
	bike,
	ped./co.},
y grid style={white!69.0196078431373!black},
ymax=25,
ymin=-3,
%ymode=log,
ytick style={color=black}
]

\path [draw=color2, semithick]
(axis cs:0.5,0.115759134292603)
--(axis cs:0.5,9.21403217315674);

\path [draw=color2, semithick]
(axis cs:1.5,1.28960084915161)
--(axis cs:1.5,14.9627795219421);

\path [draw=color2, semithick]
(axis cs:2.5,4.55585885047913)
--(axis cs:2.5,23.0810160636902);

\path [draw=color2, semithick]
(axis cs:3.5,0.0437045618891716)
--(axis cs:3.5,1.13095793128014);

\path [draw=color2, semithick]
(axis cs:4.5,0.16365122795105)
--(axis cs:4.5,3.83924758434296);

\path [draw=color2, semithick]
(axis cs:5.5,0.192202873528004)
--(axis cs:5.5,1.27387450635433);

\addplot [semithick, color2, mark=-, mark size=3, mark options={solid}, only marks]
table {%
	0.5 0.115759134292603
	1.5 1.28960084915161
	2.5 4.55585885047913
	3.5 0.0437045618891716
	4.5 0.16365122795105
	5.5 0.192202873528004
};
\addplot [semithick, color2, mark=-, mark size=3, mark options={solid}, only marks]
table {%
	0.5 9.21403217315674
	1.5 14.9627795219421
	2.5 23.0810160636902
	3.5 1.13095793128014
	4.5 3.83924758434296
	5.5 1.27387450635433
};
\addplot [semithick, color2, only marks, mark options={fill=color2, draw=color2, scale=0.75}]
table {%
	0.5 4.53738141059875
	1.5 6.45354843139648
	2.5 10.3928246498108
	3.5 0.543366551399231
	4.5 1.83687525987625
	5.5 0.601538062095642
};
\end{axis}

\vspace{10pt}
\end{tikzpicture}

%% file: figures/class_switches_nofade.tex
% This file was created with tikzplotlib v0.9.12.
\begin{tikzpicture}

\definecolor{color0}{rgb}{0.2823529411764706, 0.23921568627450981, 0.54509892156862}
\definecolor{color1}{rgb}{1.0, 0.5725490196078431, 0.1411764705882353}

\begin{axis}[
height=151.86214363138544,
tick align=outside,
tick pos=left,
width=235.71811/2,
x grid style={white!69.0196078431373!black},
xmin=-0.5999999959021806, xmax=4.50000024996698,
xlabel={Decoder layer},
ylabel={No. of class switches},
xtick style={color=black},
xtick={0,1,2,3,4},
xticklabels={0,1,2,3,4},
y grid style={white!69.0196078431373!black},
ymax=84,
ymin=-3,
%ymode=log,
ytick style={color=black}
]

\path [draw=color1, semithick]
(axis cs:0,5)
--(axis cs:0,82);

\path [draw=color1, semithick]
(axis cs:1,4)
--(axis cs:1,44);

\path [draw=color1, semithick]
(axis cs:2,3)
--(axis cs:2,34);

\path [draw=color1, semithick]
(axis cs:3,2)
--(axis cs:3,21);

\path [draw=color1, semithick]
(axis cs:4,2)
--(axis cs:4,9);

\addplot [semithick, color1, mark=-, mark size=3, mark options={solid}, only marks]
table {%
	0 5
	1 4
	2 3
	3 2
	4 2
};
\addplot [semithick, color1, mark=-, mark size=3, mark options={solid}, only marks]
table {%
	0 82
	1 44
	2 34
	3 21
	4 9
};
\addplot [semithick, color1]
table {%
	0 38
	1 20
	2 15
	3 9
	4 4
};
\addplot [name path=upper0b,draw=none] table{%
	0 82
1 44
2 34
3 21
4 9
};
\addplot [name path=lower0b,draw=none] table{%
	0 5
1 4
2 3
3 2
4 2
};
\addplot [fill=color1!10] fill between[of=upper0b and lower0b];

\end{axis}
\vspace{10pt}
\end{tikzpicture}

%% file: figures/class_switches_fade.tex
% This file was created with tikzplotlib v0.9.12.
\begin{tikzpicture}

\definecolor{color0}{rgb}{0.2823529411764706, 0.23921568627450981, 0.54509892156862}
\definecolor{color1}{rgb}{1.0, 0.5725490196078431, 0.1411764705882353}

\begin{axis}[
height=151.86214363138544,
tick align=outside,
tick pos=left,
width=235.71811/2,
x grid style={white!69.0196078431373!black},
xmin=-0.5999999959021806, xmax=4.50000024996698,
xlabel={Decoder layer},
ylabel={No. of class switches},
xtick style={color=black},
xtick={0,1,2,3,4},
xticklabels={0,1,2,3,4},
y grid style={white!69.0196078431373!black},
ymax=84,
ymin=-3,
%ymode=log,
ytick style={color=black}
]
\path [draw=color0, semithick]
(axis cs:0,6)
--(axis cs:0,41);

\path [draw=color0, semithick]
(axis cs:1,4)
--(axis cs:1,22);

\path [draw=color0, semithick]
(axis cs:2,3)
--(axis cs:2,19);

\path [draw=color0, semithick]
(axis cs:3,2)
--(axis cs:3,13);

\path [draw=color0, semithick]
(axis cs:4,1)
--(axis cs:4,5);

\addplot [semithick, color0, mark=-, mark size=3, mark options={solid}, only marks]
table {%
0 6
1 4
2 3
3 2
4 1
};
\addplot [semithick, color0, mark=-, mark size=3, mark options={solid}, only marks]
table {%
0 41
1 22
2 19
3 13
4 5
};
\addplot [semithick, color0]
table {%
0 17
1 9
2 7
3 5
4 2
};
\addplot [name path=upper0,draw=none] table{%
0 41
1 22
2 19
3 13
4 5
};
\addplot [name path=lower0,draw=none] table{%
0 6
1 4
2 3
3 2
4 1
};

\addplot [fill=color0!10] fill between[of=upper0 and lower0];

\end{axis}
\vspace{10pt}
\end{tikzpicture}